\documentclass{article}

\usepackage{arxiv}
\usepackage{graphicx}
\usepackage{bbold}
\usepackage{optidef}
\usepackage{multirow}

\usepackage{amsthm}
\theoremstyle{definition}
\newtheorem{defn}{Definition}
\theoremstyle{plain}
\newtheorem{thm}{Theorem}
\newtheorem{lem}[thm]{Lemma}

\usepackage[utf8]{inputenc}
\usepackage[T1]{fontenc}
\usepackage{optidef}
\usepackage{hyperref}
\usepackage{nicematrix}
\hypersetup{
    colorlinks=true,
    linkcolor=blue,
    filecolor=magenta,      
    urlcolor=black,
}
\usepackage[numbers]{natbib}

\usepackage[utf8]{inputenc} 
\usepackage[T1]{fontenc}    
\usepackage{hyperref}       
\usepackage{url}            
\usepackage{booktabs}       
\usepackage{amsmath}
\usepackage{amsfonts}       
\usepackage{nicefrac}       
\usepackage{microtype}      
\usepackage{lipsum}		
\usepackage{graphicx}
\usepackage{doi}

\newcommand{\argmin}{\mathop{\rm arg~min}\limits}
\title{On a linear fused Gromov-Wasserstein distance for graph structured data}

\author{ \href{https://orcid.org/0000-0003-0380-4197}{\includegraphics[scale=0.06]{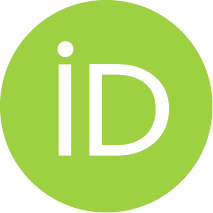}\hspace{1mm}Dai Hai Nguyen}\\
	Graduate School of Frontier Sciences\\
	The University of Tokyo\\
	5-1-5 Kashiwa-no-ha, Kashiwa, Chiba 277-8561, Japan \\
	\texttt{hai@k.u-tokyo.ac.jp} \\
	\And
	\href{https://orcid.org/0000-0002-4288-1606}{\includegraphics[scale=0.06]{orcid.pdf}\hspace{1mm}Koji Tsuda} \\
	Graduate School of Frontier Sciences\\
		The University of Tokyo\\
	5-1-5 Kashiwa-no-ha, Kashiwa, Chiba 277-8561, Japan  \\
	\texttt{tsuda@k.u-tokyo.ac.jp} \\
}

\renewcommand{\shorttitle}{\texttt{linearFGW}: a novel optimal transport metric for graph structured data}

\hypersetup{
pdftitle={A template for the arxiv style},
pdfsubject={q-bio.NC, q-bio.QM},
pdfauthor={David S.~Hippocampus, Elias D.~Striatum},
pdfkeywords={First keyword, Second keyword, More},
}

\begin{document}
\maketitle

\begin{abstract}
We present a framework for embedding graph structured data into a vector space, taking into account node features and topology of a  graph into the optimal transport (OT) problem. Then we propose a novel distance between two graphs, named \texttt{linearFGW}, defined as the Euclidean distance between their embeddings. The advantages of the proposed distance are twofold: 1) it can take into account node feature and structure of graphs for measuring the similarity between graphs in a kernel-based framework, 2) it can be much faster for computing kernel matrix than pairwise OT-based distances, particularly fused Gromov-Wasserstein, making it possible to deal with large-scale data sets. After discussing theoretical properties of \texttt{linearFGW}, we demonstrate experimental results on classification and clustering tasks, showing the effectiveness of the proposed \texttt{linearFGW}.
\end{abstract}


\section{Introduction}
Many applications of machine learning involve learning with graph structured data such as bioinformatics \cite{sharan2006modeling}, social networks \cite{scott2011social}, chemoinformatics \cite{trinajstic2018chemical}, and so on. To deal with graph structured data, many graph kernels have been proposed in literature for measuring the similarity between graphs in a kernel-based framework. Most of them are based on $\mathcal{R}$-framework, which focuses on comparing graphs based on their substructures such as subtree \cite{shervashidze2011weisfeiler}, shortest path \cite{borgwardt2005shortest}, random walk \cite{kashima2003marginalized}, and so on. However, these methods have several limitations: 1) they do not consider feature and structure distributions of graphs, 2) they require to define substructures based on the domain knowledge, which might not be available in many practical applications.

Optimal Transport (OT) \cite{villani2009wasserstein} has received much attention in the machine learning community and has been shown to be an effective tool for comparing probability measures in many applications. In recent years, several studies have attempted to use the OT distance for learning graph structured data by considering the problem of measuring the similarity of graphs as an instance of computing OT distance for graphs. Togninalli et al. \cite{togninalli2019wasserstein} introduced the Wasserstein distance to compare graphs based on their node embeddings obtained by Weisfeiler-Lehmann labeling framework \cite{shervashidze2011weisfeiler}. Titouan et al. \cite{titouan2019optimal} proposed fused Gromov-Wasserstein (\texttt{FGW}) which combines Wasserstein and Gromov-Wasserstein \cite{memoli2011gromov,peyre2016gromov} distances in order to jointly take into account features and structures of graphs. These OT-based distances have achieved great performance for graph classification. However, they have several limitations: 1) kernel matrices converted from the OT-based distances are generally not valid, so they are not ready to use for the kernel-based framework, 2) calculating the similarity between each pair of graphs is computationally expensive, so the need for computing the kernel matrix of all pairwise similarity can be a burden for dealing with large-scale graph data sets.

In order to overcome the aforementioned limitations, inspired by the linear optimal transport framework introduced by Wang et al. \cite{wang2013linear}, we propose an OT-based distance, named \texttt{linearFGW}, for learning with graph structured data. As the name suggests, our distance is a generalization of the linear optimal transport framework and \texttt{FGW} distance. The basic idea is to embed the node features and topology of a graph into a linear tangent space through a fixed reference measure graph. Then the \texttt{linearFGW} distance between two graphs is defined as the Euclidean distance between their two embeddings, which approximates their \texttt{FGW} distance. Therefore, the \texttt{linearFGW} distance has the following advantages: 1) it can take into account node features and topologiesb of graphs for the OT problem in order to calculate the dissimilarity between graphs, 2) we can derive a valid graph kernel from the embeddings of graphs for the downstream tasks such as graph classification and clustering, 3) by using the \texttt{linearFGW} as an approximate of the \texttt{FGW}, we can avoid expensive computation of the pairwise \texttt{FGW} distance for large-scale graph data sets. Finally, we conduct experiments on graph data sets to show the effectiveness of the proposed distance in terms of classification and clustering accuracies.

The remainder of the paper is organised as follows: in Section 2, we present some related work. In Section 3, we present the idea of our proposed distance for learning graph with structured data and its theoretical properties. In Section 4, experimental results on benchmark graph data sets are provided. Finally, we conclude by summarizing this work and discussing possible extensions in Section 5.
\section{Related Work}
\subsection{Kernels for Graphs}
Graph is a standard representation for relational data, which appear in various domains such as bioinformatics \cite{sharan2006modeling}, chemoinformatics \cite{trinajstic2018chemical}, social network analysis \cite{scott2011social}. Making use of graph kernels is a popular approach to learning with graph structured data. Essentially, a graph kernel is a measure of the similarity between two graphs and must satisfy two fundamental requirements to be a valid kernel: 1) symmetric and 2) positive semi-definite (PSD). There are a number of kernels for graphs with discrete attributes such as random walk \cite{kashima2003marginalized}, shortest path \cite{kashima2003marginalized}, Weisfeiler-Lehman (WL) subtree \cite{shervashidze2011weisfeiler} kernels, just to name a few. There are several kernels for graphs with continuous attributes such as GraphHopper \cite{feragen2013scalable}, Hash Graph \cite{morris2016faster} kernels.
\subsection{Optimal Transport Frameworks for Graphs}
Optimal Transport (OT) \cite{villani2009wasserstein} has received much attention from the machine learning community as it provide an effective way to measure the distance between two probability measures. Some OT-based graph kernels have been proposed and achieved great performance in comparison with traditional graph kernels. Wasserstein Weisfeiler-Lehman (WWL) \cite{togninalli2019wasserstein} used OT for measuring distance between two graphs based on their WL embeddings (discrete feature vectors of subtree patterns). Nguyen et al. \cite{nguyen2021learning} extends WWL by proposing an efficient algorithm for learning subtree pattern importance, leading to higher classification accuracy on graph data sets. However, they are not valid kernels for graphs with continuous attributes. Following the work in \cite{memoli2011gromov}, Peyre et al. \cite{peyre2016gromov} proposed a Gromow-Wasserstein distance to compare pairwise similarity matrices from different spaces. Then, Titouan et al. \cite{titouan2019optimal} proposed a fused Gromov-Wasserstein distance which combine Wasserstein and Gromov-Wasserstein distances in order to jointly leverage feature and structure information of graphs. To reduce computational complexity, OT-based distances are often computed using a Sinkhorn algorithm \cite{sinkhorn1967diagonal, cuturi2013sinkhorn}. Due to the nature of optimal assignment problem, these OT-based graph kernels are indefinite similarity matrices so they are invalid kernels, leading to the use of support vector machine (SVM) with indefinite kernels introduced in \cite{luss2007support}.

\subsection{Linear Optimal Transport}
Wang et al. \cite{wang2013linear} proposed a simplified version of OT in 2-Wasserstein space, called linear optimal transport. In the sense of geometry, the basic idea is to transfer probability measures from the geodesic 2-Wasserstein space to the tangent space with respect to some fixed base or reference measure. One advantage is that we can work on a linear tangent space instead of the complex 2-Wasserstein space so that the downstream tasks such as classification, clustering can be done in the linear space. Another advantage is the fast approximation of pairwise Wasserstein distance for large-scale data sets. In the context of graph learning, Kolouri et al. \cite{kolouri2020wasserstein} leveraged the above framework and introduced the concept of linear Wasserstein embedding for learning graph embeddings. In a concurrent work, Mialon et al. \cite{mialon2020trainable} proposed a similar idea for learning set of features. In this paper, we extend the idea of linear optimal transport framework from the 2-Wasserstein distance to the Fused Gromov-Wasserstein distance (\texttt{FGW}), and define a valid graph kernel for learning with graph structured data. Furthermore, we derive theoretical understandings of the proposed distance.

\section{Proposed Distance for Graphs: Linear Fused Gromov-Wasserstein}
We denote a measure graph as $\mathcal{G}(\mathbf{X}, \mathbf{A}, \mu)$, where $\mathbf{X}=\{\mathbf{x}_{i}\}_{i=1}^{m}\in \mathbb{R}^{m\times d}$ is the set of $m$ node features with dimensionality of $d$, $\mathbf{A}=\left[a\right]_{ij}\in \mathbb{R}^{m \times m}$ is a square matrix to encode the topology of the given graph such as the adjacency matrix or the matrix of pairwise distances between nodes, $\mu=\left[\mu_{i} \right]\in \Delta^{m}$ (probability simplex) is a Borel probability measure defined on the nodes (note that when no additional information is provided,  all probability measures can be set as uniform). 

\subsection{\texttt{FGW}: A Distance for Matching Node Features and Structures}
\label{subsection:fgw}
In \cite{titouan2019optimal}, a graph distance, named Fused Gromov-Wasserstein (\texttt{FGW}), is proposed to take into account both node feature and topology information into the OT problem for measuring the dissimilarity between two graphs. Formally, given two graphs $\mathcal{G}_{1}(\mathbf{X}, \mathbf{A}, \mu)$ and $\mathcal{G}_{2}(\mathbf{Y}, \mathbf{B}, \nu)$, the \texttt{FGW} distance between $\mathcal{G}_{1}$ and $\mathcal{G}_{2}$ is defined for a trade-off parameter $\alpha\in \left[0,1 \right]$ as:
\begin{align}
\label{eqn:fgw}
    \texttt{FGW}_{q,\alpha}(\mathcal{G}_{1}, \mathcal{G}_{2})=\min_{\pi \in \Pi(\mu,\nu)}\sum_{i,j,k,l}\left[(1-\alpha)\lVert \mathbf{x}_{i}-\mathbf{y}_{j}\rVert^{q}+\alpha |\mathbf{A}_{i,k}-\mathbf{B}_{j,l}|^{q}\right]\pi_{i,j}\pi_{k,l}
\end{align}
where $\Pi(\mu,\nu)=\{\pi\in R_{+}^{m\times n} \text{s.t. }\sum_{i=1}^{m}\pi_{i,j}=\nu_{j}\text{, } \sum_{j=1}^{n}\pi_{i,j}=\mu_{i} \}$ is the set of all admissible couplings between $\mu$ and $\nu$. The \texttt{FGW} distance acts as a generalization of the Wasserstein \cite{villani2009wasserstein} and Gromov-Wasserstein \cite{memoli2011gromov}, which allows balancing the importance of matching the node features and topologies between two graphs. However, similar to the existing OT-based graph distances, it is challenging to define a valid kernel from the \texttt{FGW} for the graph-related prediction task, due to the nature of optimal assignment problem. In the following, we restrict our attention to the OT with $q=2$ and for the ease of presentation, we use the notation $\texttt{FGW}_{\alpha}$ instead of $\texttt{FGW}_{q,\alpha}$.

\subsection{\texttt{linearFGW}: A New Distance for Comparing Graphs}
\label{subsection:linearFGW}
\begin{figure}[t]
	\centerline{\includegraphics[width=0.9\columnwidth]{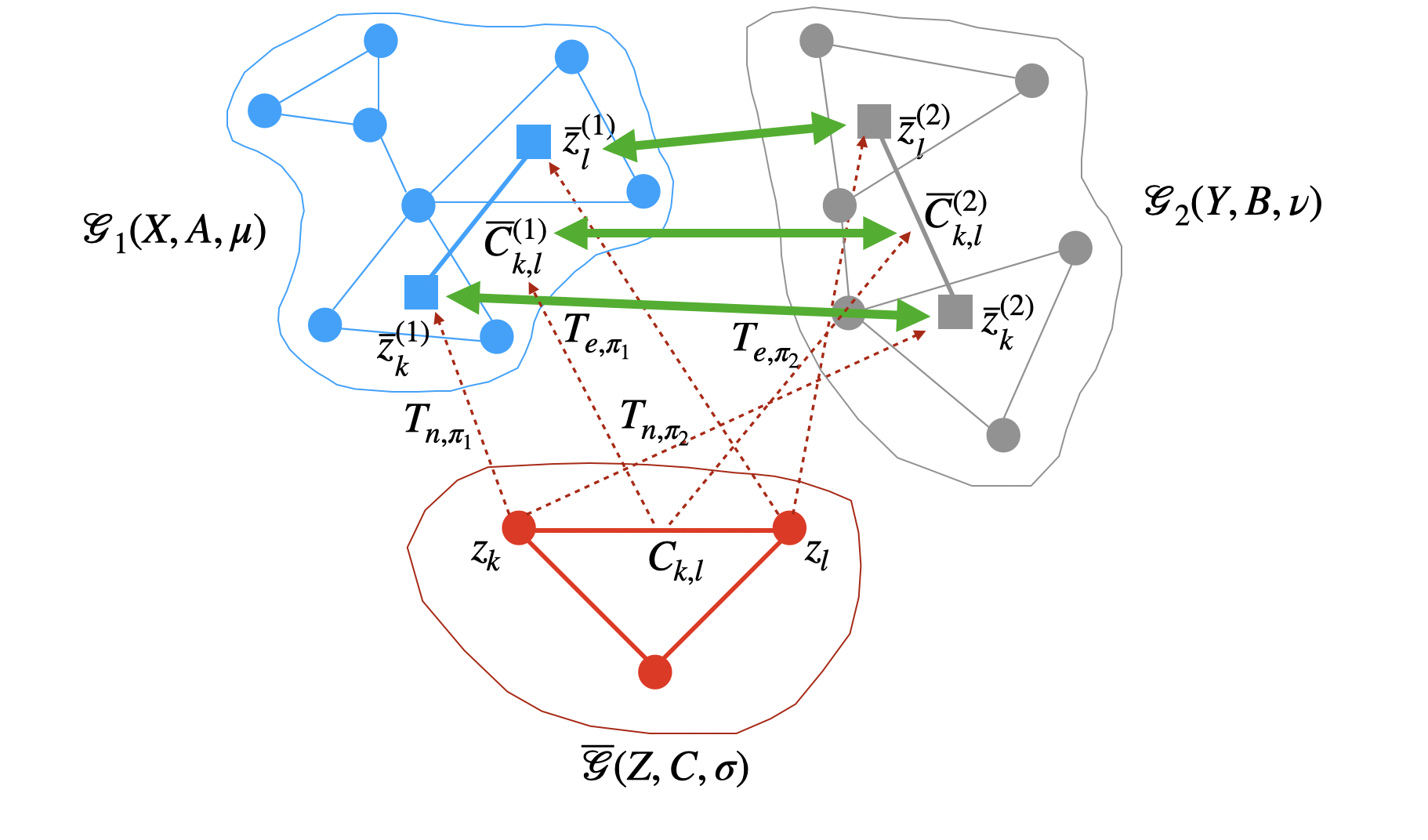}}
	\caption{Illustration of the computation of the \texttt{linearFGW} distance between $\mathcal{G}_{1}(\mathbf{X},\mathbf{A},\mu)$ and $\mathcal{G}_{2}(\mathbf{Y},\mathbf{B},\nu)$, given the fixed reference measure graph $\overline{\mathcal{G}}(\mathbf{Z},\mathbf{C},\sigma)$. First, we find the optimal transport plans $\pi_{1}$ and $\pi_{2}$ from $\overline{\mathcal{G}}$ to $\mathcal{G}_{1}$ and $\mathcal{G}_{2}$, respectively, in the sense of the \texttt{FGW}. Then we transport $\overline{\mathcal{G}}$ with the barycentric projections for nodes and edges (see Definition  \ref{def:barycentricprojections}) using the optimal plans $\pi_{1}$ and $\pi_{2}$ to get the surrogate measure graphs $\Tilde{\mathcal{G}}_{1}(\Tilde{\mathbf{Z}}^{(1)}, \Tilde{\mathbf{C}}^{(1)}, \sigma)$ and $\Tilde{\mathcal{G}}_{2}(\Tilde{\mathbf{Z}}^{(2)}, \Tilde{\mathbf{C}}^{(2)}, \sigma)$ for $\mathcal{G}_{1}$ and $\mathcal{G}_{2}$, respectively. Finally, the Euclidean distance between $\Tilde{\mathcal{G}}_{1}$ and $\Tilde{\mathcal{G}}_{2}$ can be directly calculated using Equation (\ref{eqn:linearfgw}).}
	\label{fig:linearfgw}
\end{figure}

In order to overcome the limitations of the \texttt{FGW} distance, we propose to approximate it by a linear optimal transport framework, which we call Linear Fused Gromov-Wasserstein (\texttt{LinearFGW}). Computing the \texttt{LinearFGW} distance requires a reference, which we choose to be also a measure graph $\overline{\mathcal{G}}(\mathbf{Z}, \mathbf{C}, \sigma)$. How the reference measure graph is chosen is described later in Subsection \ref{subsection:reference}. 
To precisely define the \texttt{LinearFGW} distance, we first define the barycentric projections for node features and structures of graphs as follows:
\begin{defn}[Barycentric projections for nodes and edges of graphs]
\label{def:barycentricprojections}
Let $\overline{\mathcal{G}}(\mathbf{Z}, \mathbf{C}, \sigma)$ be a measure graph with $K$ nodes. Denote by
\[
\pi = \sum_{k=1}^{K}\sum_{i=1}^{m} \pi_{k,i}\, \delta_{(\mathbf{z}_k, \mathbf{x}_i)}
\]
a transport plan from $\overline{\mathcal{G}}$ to $\mathcal{G}$, i.e.\ a coupling of $\sigma$ and $\mu$. The \emph{barycentric projections} of the nodes and edges of $\overline{\mathcal{G}}$ onto the space of $\mathcal{G}$ induced by $\pi$ are defined, for $k,l \in \{1,\dots,K\}$, by
\begin{equation}
\label{eqn:barycentricprojections}
T_{\mathrm{n},\pi}(\mathbf{z}_k) \;=\; \frac{1}{\sigma_k}\sum_{i=1}^{m} \pi_{k,i}\, \mathbf{x}_i,
\qquad
T_{\mathrm{e},\pi}(\mathbf{C}_{k,l}) \;=\; \frac{1}{\sigma_k \sigma_l}\sum_{i=1}^{m}\sum_{j=1}^{m} \pi_{k,i}\, \pi_{l,j}\, \mathbf{A}_{i,j}.
\end{equation}
\end{defn}
  The definitions of these projections are extended from \cite{wang2013linear,beier2021linear}. Furthermore, we derive their properties in the following lemma.
\begin{lem}
\label{lemma:optimalplan} Given two measure graphs $\overline{\mathcal{G}}(\mathbf{Z}, \mathbf{C}, \sigma)$ and $\mathcal{G}(\mathbf{X}, \mathbf{A}, \mu)$, we denote $\pi^{*}$ as the optimal transport plan from $\overline{\mathcal{G}}$ to $\mathcal{G}$ with respect to the \texttt{FGW} distance, and $\Tilde{\mathcal{G}}(\Tilde{\mathbf{Z}},\Tilde{\mathbf{C}},\sigma)$ as the probability measure graph obtained by applying barycentric projections for nodes and edges $T_{\text{n}, \pi^{*}}(.)$ and $T_{\text{e}, \pi^{*}}(.)$, respectively (see Definition \ref{def:barycentricprojections}). Then, we have the following claims:
\begin{enumerate}
    \item $\operatorname{diag}(\sigma)=\begin{bmatrix}\sigma_{1} & 0 & 0\\0 & \ddots & 0\\ 0 & 0 & \sigma_{K}\end{bmatrix}$ is the optimal transport plan from $\overline{\mathcal{G}}$ to $\Tilde{\mathcal{G}}$ in the sense of the \texttt{FGW} distance.
    \item $\texttt{FGW}_{\alpha}(\overline{\mathcal{G}}, \Tilde{\mathcal{G}})\leq \texttt{FGW}_{\alpha}(\overline{\mathcal{G}}, \mathcal{G})$.
\end{enumerate}
\end{lem}
The proof is given in the Appendix section. An important implication of the above lemma is that
$\Tilde{\mathcal{G}}$ can be considered as a surrogate measure graph for $\mathcal{G}$ with respect to the reference $\overline{\mathcal{G}}$. Thus we propose to define the \texttt{LinearFGW} distance between two  measure graphs $\mathcal{G}_{1}$ and $\mathcal{G}_{2}$ with respect to the reference measure graph $\overline{\mathcal{G}}$ as follows:
\begin{align}
\label{eqn:linearfgw}
   \texttt{linearFGW}_{\alpha}(\mathcal{G}_{1}, \mathcal{G}_{2})=(1-\alpha) \sum_{k}\lVert T_{\text{n}, \pi_{1}}(\mathbf{z}_{k})-T_{\text{n}, \pi_{2}}(\mathbf{z}_{k})\rVert^{2} + \alpha \sum_{k,l}|T_{\text{e}, \pi_{1}}(\mathbf{C}_{k,l})-T_{\text{e}, \pi_{2}}(\mathbf{C}_{k,l})|^{2}
\end{align}
where $\pi_{1}$ and $\pi_{2}$ denote the optimal transport plans from $\overline{\mathcal{G}}$ to $\mathcal{G}_{1}$ and $\mathcal{G}_{2}$, respectively, in the sense of the \texttt{FGW} distance. We call this distance \texttt{linearFGW} as it acts as a generalization of linear optimal transport \cite{wang2013linear} and \texttt{FGW} \cite{titouan2019optimal}. Furthermore, the proposed distance also suggests a Euclidean embedding of the measure graph $\mathcal{G}_{1}$ with respect to the reference measure graph $\overline{\mathcal{G}}$ becomes:
$\Phi_{\overline{\mathcal{G}}, \alpha}(\mathcal{G}_{1})=\left(\sqrt{1-\alpha}T_{\text{n}, \pi_{1}}(\mathbf{z}_{1}),...,\sqrt{1-\alpha}T_{\text{n}, \pi_{1}}(\mathbf{z}_{K}),...,\sqrt{\alpha}T_{\text{e}, \pi_{1}}(\mathbf{C}_{k,l}),...\right)$ with dimension of $K+K^{2}$. So we can derive a valid kernel for graph-related prediction tasks. The computation of the \texttt{linearFGW} can be illustrated in Figure \ref{fig:linearfgw}.

\subsection{Selection of Reference Measure Graph}
\label{subsection:reference}
Selecting the reference measure graph in Subsection \ref{subsection:linearFGW}  is important. We empirically observe that if the reference is randomly selected or distant from all measure graphs, the approximation error between \texttt{FGW} and \texttt{linearFGW} is likely to increase. In the lemma presented below, we show the relation between \texttt{FGW} and \texttt{linearFGW} with respect to the reference measure graph.

\begin{lem}
\label{Lemma2} We denote the mixing diameter of a graph $\mathcal{G}(\mathbf{X}, \mathbf{A}, \mu)$  by $\texttt{diam}_{\alpha}(\mathcal{G})=\alpha \max_{i,j}\lVert \mathbf{x}_{i}-\mathbf{x}_{j} \rVert^{2} + (1-\alpha)\max_{i,j,i^\prime, j^\prime}|\mathbf{A}_{i,j}-\mathbf{A}_{i^\prime, j^\prime}|^{2}$. Then, given a fixed reference measure graph $\overline{\mathcal{G}}(\mathbf{Z}, \mathbf{C}, \sigma)$,
for two input measure graphs $\mathcal{G}_{1}(\mathbf{X}, \mathbf{A}, \mu)$ and $\mathcal{G}_{2}(\mathbf{Y}, \mathbf{B}, \nu)$, we have the following inequality:
\begin{equation}
\label{lemma:barycentric}
    |\texttt{FGW}_{\alpha}(\mathcal{G}_{1}, \mathcal{G}_{2})-\texttt{linearFGW}_{\alpha}(\mathcal{G}_{1}, \mathcal{G}_{2})|\leq 4\min\{\texttt{FGW}_{\alpha}(\mathcal{G}_{1}, \overline{\mathcal{G}}),\texttt{FGW}_{\alpha}(\mathcal{G}_{2}, \overline{\mathcal{G}})\} + 2\texttt{diam}_{\alpha}(\mathcal{G}_{1}) +  2\texttt{diam}_{\alpha}(\mathcal{G}_{2})
\end{equation}
\end{lem}

The proof is given in the Appendix section. 

A corollary of the above lemma suggests how to select a good reference measure graph $\overline{\mathcal{G}}$: given $N$ graphs $(\mathcal{G}_{1},...,\mathcal{G}_{N})$, the total approximation error is upper bounded by:
\begin{equation}
    \sum_{i=1}^{N}\sum_{j=i+1}^{N}|\texttt{FGW}_{\alpha}(\mathcal{G}_{i}, \mathcal{G}_{j})-\texttt{linearFGW}_{\alpha}(\mathcal{G}_{i}, \mathcal{G}_{j})|\leq 4\sum_{i=1}^{N}\text{FGW}_{\alpha}(\mathcal{G}_{i}, \overline{\mathcal{G}}) + 4\sum_{i=1}^{N}\texttt{diam}_{\alpha}(\mathcal{G}_{i})
\end{equation}
where the right-hand side has two terms: the first term is the objective of the fused Gromov-Wasserstein barycenter problem \cite{titouan2019optimal} while the second term is constant with respect to the reference measure graph $\overline{\mathcal{G}}$, suggesting that we can use the fused Gromov-Wasserstein barycenter of $N$ given measure graphs as the reference.
\subsection{Implementation Details}
The \texttt{FGW} is the main component of our method. We use the proximal point algorithm (PPA) \cite{xu2020gromov} to implement the \texttt{FGW}. Specifically, given two graphs $\mathcal{G}_{1}$ and $\mathcal{G}_{2}$, we solve the problem (\ref{eqn:fgw}) iteratively ( with maximum $T$ iterations) as follows:
\begin{align}
    \pi^{(t+1)}=\argmin_{\pi \in \Pi(\mu,\nu)}\langle (1-\alpha)\mathbf{D}_{12} + \alpha (\mathbf{C}_{12}-2 \mathbf{A}\pi^{(t)}\mathbf{B}), \pi \rangle + \eta \textbf{KL}(\pi|| \pi^{(t)})
    \label{eqn:sinkhorniterations}
\end{align}
where $\langle\cdot,\cdot\rangle$ denote the inner product of matrices, $\mathbf{D}_{12}=(\mathbf{X}\odot \mathbf{X})\mathbf{1}_{d}\mathbf{1}_{n}^\top+\mathbf{1}_{m}\mathbf{1}_{d}^\top(\mathbf{Y}\odot \mathbf{Y})^\top$, $\mathbf{C}_{12}=(\mathbf{A}\odot \mathbf{A})\mu \mathbf{1}_{n}^\top+\mathbf{1}_{m}\nu^\top(\mathbf{B}\odot \mathbf{B})^\top$ and $\odot$ denotes the Hadamard product of matrices. $\textbf{KL}(\pi|| \pi^{(t)})$ is the Kullback-Leibler divergence between the optimal transport plan and the previous estimation. We can approximately solve the above problem by Sinkhorn-Knopp update (see \cite{xu2020gromov} for the algorithmic details).
\begin{table}[]
    \centering
    \caption{Statistics of data sets used in experiments}
    \begin{tabular}{c|c c c c c}
    \hline\hline
    Dataset & \#graphs & \#classes & Ave. \#odes & Ave. \#edges & \#attributes\\
    \hline
    COX2 & 467 &  2 & 41.22 & 43.45 & 3\\
   BZR & 405 &  2 & 35.75 & 38.36 & 3\\
    ENZYMES & 600 &  6 & 32.63 & 62.14 & 18\\
   PROTEINS & 1113 & 2 & 39.06 & 72.82 & 1\\
    PROTEINS-F & 1113 & 2 & 39.06 & 72.82 & 29\\
     AIDS & 2000 & 2 & 15.69 & 16.20 & 4\\
     IMDB-B & 1000 & 2 & 19.77 & 96.53 & -\\
    \hline\hline
    \end{tabular}
    \label{tab:datasets}
\end{table}
\section{Experimental Results}
We now show the effectiveness of our proposed graph distance 
on real world data sets in terms of  graph classification and clustering. Our code can be accessed via the following link: \texttt{https://github.com/haidnguyen0909/linearFGW}.
\subsection{Data sets}
In this work, we focus on graph kernels/distances for graphs with continuous attributes. So we consider the following seven widely used benchmark data sets: BZR \cite{sutherland2003spline}, COX2 \cite{sutherland2003spline}, ENZYMES \cite{dobson2003distinguishing}, PROTEINS \cite{borgwardt2005protein}, PROTEINS-F \cite{borgwardt2005protein}, AIDS \cite{riesen2008iam} contain graphs with continous attributes, while IMDB-B \cite{yanardag2015deep} contains unlabeled graphs obtained from social networks. All these data sets can be downloaded from \texttt{https://ls11-www.cs.tu-dortmund.de/staff/morris/graphkerneldatasets}. The details of the used data sets are shown in Table \ref{tab:datasets}.

\subsection{Experimental settings}
To compute numerical features for nodes of graphs, we consider two main settings: 1) we keep the original attributes of nodes (denoted by suffix \texttt{RAW}), 2) we consider Weisfeiler-Lehman (WL) mechanism by concatenating numerical vectors of neighboring nodes (denoted by the suffix $\texttt{WL}-$H where H means we repeat the procedure H times to take neighboring vertices within H hops into account to obtain the features, see \cite{shervashidze2011weisfeiler} for more detail). For the matrix $\textbf{A}$, we restrict our attention to the adjacency matrices of the input graphs. For solving the optimization problem (\ref{eqn:sinkhorniterations}), we fix $\eta$ as 0.1 and the number of iterations $T$ as 5. We carry out our experiments on a 2.4 GHz 8-Core Intel Core i9 with 64GB RAM.

For the classification task, we convert a distance into a kernel matrix through the exponential function, i.e. $K=\exp(-\gamma D)$ (Gaussian kernel). We compare the classification accuracy with the following state-of-the-art graph kernels (or distances): GraphHopper kernel (GH, \cite{feragen2013scalable}), HGK-WL \cite{morris2016faster}, HGK-SP \cite{morris2016faster}, RBF-WL \cite{togninalli2019wasserstein}, Wasserstein Weisferler-Lehman kernel (WWL, \cite{togninalli2019wasserstein}), FGW \cite{titouan2019optimal}, GWF \cite{xu2020gromov}. We divide them into two groups: OT-based graph kernels including WWL, FGW, GFW and \texttt{linearFGW} (ours) and non-OT graph kernels including GH, HGK-WL, HGK-SP, RBF-WL. Note that our proposed graph kernel converted from the \texttt{linearFGW} is the only (valid) positive definite kernel among OT-based graph kernels. 

We perform 10-fold cross validation and report the average accuracy of the experiment repeated 10 times. The accuracies of other graph kernels are taken from the original papers. We use SVM for classification and cross-validate the parameters $C=\{2^{-5},2^{-4},...,2^{10}\}$, $\gamma=\{10^{-2},10^{-1},...,10^{2}\}$. The range of the WL parameter $H={1,2}$. For our proposed \texttt{linearFGW}, $\alpha$ is cross-validated via a search in $\{0.0,0.3,0.5,0.7,0.9,1.0\}$. Note that the linear optimal transport \cite{wang2013linear} is a special case of the \texttt{linearFGW} with $\alpha=0$.

We also compare the clustering accuracy with OT-based graph distances: \texttt{FGW}, GWB-KM, GWF on four real world data sets: AIDS, PROTEINS, PROTEINS-F, IMDB-B. For fair comparison, we use K-means and spectral clustering on the Euclidean embedding and Gaussian kernel of the proposed \texttt{linearFGW} distance (denoted by \texttt{linearFGW-Kmeans} and \texttt{linearFGW-SC}, respectively). We fix the parameters $H=1$, $\alpha=0.5$ for data sets of graphs with continuous attributes and $\gamma=0.01$ for the Gaussian kernel.

\begin{table}[]
    \centering
    \caption{Average classification accuracy on the graph data sets with vector attributes. The best result for each column (data set) is highlighted in bold and the standard deviation is reported with the symbol $\pm$.}
    \begin{tabular}{l l l l l l l}
    \hline\hline
        & Kernels/Data sets & COX2 & BZR & ENZYMES & PROTEINS & IMDB-B\\
    \hline
    \multirow{4}{*}{Non OT} & GH & 76.41$\pm$ 1.39 & 76.49$\pm$ 0.9 & 65.65$\pm$ 0.8 & 74.48$\pm$ 0.3 & -\\
                                  & HGK-WL & 78.13$\pm$ 0.45 & 78.59$\pm$ 0.63 & 63.04$\pm$ 0.65 & 75.93$\pm$ 0.17 & -\\
                                  & HGK-SP & 72.57$\pm$ 1.18 & 76.42$\pm$ 0.72 & 66.36$\pm$ 0.37 & 75.78$\pm$ 0.17 & -\\
                                  & RBF-WL & 75.45$\pm$ 1.53 & 80.96$\pm$ 1.67 & 68.43$\pm$ 1.47 & 75.43$\pm$ 0.28 & -\\
    \hline
     \multirow{4}{*}{OT-based} & WWL & 78.29$\pm$0.47 & 84.42$\pm$ 2.03 & 73.25$\pm$ 0.87 & 77.91$\pm$ 0.8 & -\\
                                  & \texttt{FGW} & 77.2$\pm$4.7 & 84.1 $\pm$4.1 & 71.0$\pm$ 6.7 & 75.1$\pm$ 2.9 & \textbf{64.2}$\pm$ 3.3\\
                                  & \texttt{GWF} & - & - & - & 73.7$\pm$2.0 & 63.9$\pm$2.7\\
                                  & \texttt{linearFGW}-RAW (Ours)  & 79.74 $\pm$ 1.99 & \textbf{86.07}$\pm$1.64 & 83.25 $\pm$ 2.44 & 82.49$\pm$ 1.75 & 63.62$\pm$1.9\\
                                  & \texttt{linearFGW}-WL1 (Ours)  & \textbf{79.98} $\pm$ 3.21 & 84.80$\pm$2.95 & \textbf{85.28}$\pm$1.64 & 83.29 $\pm$ 1.63 & -\\
                                  & \texttt{linearFGW}-WL2 (Ours) & 79.50$\pm$3.29 & 84.37$\pm$2.75 & 83.13$\pm$ 1.56 & \textbf{83.95} $\pm$ 1.12 & -\\
    \hline\hline
    \end{tabular}
    
    \label{tab:classificationresults}
\end{table}

\subsection{Results}
\textbf{Classification:} The average classification accuracies shown in Table \ref{tab:classificationresults} indicate that the \texttt{linearFGW} is a clear state-of-the-art method for graph classification. It achieved the best performances on 4 out of 6 data sets. In particular, on two data sets ENZYMES and PROTEINS, the \texttt{linearFGW} outperformed all the rest by large margins (around 12\% and 6\%, respectively) in comparison with the second best ones. On COX2 and BZR, the \texttt{linearFGW} achieved improvements of around 2\% and 1.5\%, respectively, over WWL which is the second best one. Note that the Gaussian kernel derived from WWL is not valid for the data sets of graphs with continuous attributes (see \cite{togninalli2019wasserstein}). On IMDB-B, the average accuracies of the compared methods are comparable. Interestingly, despite that our \texttt{linearFGW} is an approximate of the \texttt{FGW} distance, the \texttt{linearFGW} consistently achieved significantly higher performance than \texttt{FGW}. This can be explained by the fact that the kernel derived from the \texttt{linearFGW} distance is valid.

\textbf{Clustering:} The average clustering accuracies shown in Table \ref{tab:clusteringresults} also indicate that the \texttt{linearFGW} could achieve high performance on clustering. On PROTEINS and PROTEINS-F, the \texttt{linearFGW} achieved the highest accuracies by margins of around 2\% and 3\%, respectively, over the second best one. On AIDS and IMDB-B, the \texttt{linearFGW} achieved comparable performances with GWF-PPA which is the best performer.

\begin{table}[]
    \centering
    \caption{Average clustering accuracy on the graph data sets with continuous attributes. The best result for each column (data set) is highlighted in bold and the standard deviation is reported with the symbol $\pm$.}
    \begin{tabular}{l l l l l}
    \hline\hline
    Methods/Data sets & AIDS & PROTEINS & PROTEINS-F & IMDB-B \\
    \hline
    FGW & 91.0 $\pm$0.7 &  66.4$\pm$0.8 & 66.0$\pm$0.9 & 56.7$\pm$1.5\\
   GWB-KM & 95.2$\pm$0.9 & 64.7$\pm$1.1 & 62.9$\pm$1.3 & 53.5$\pm$2.3\\
   GWF-BADMM & 97.6$\pm$0.8 & 69.2$\pm$1.0 & 68.1$\pm$1.1 & 55.9$\pm$1.8\\
   GWF-PPA & \textbf{99.5}$\pm$0.4 & 70.7$\pm$0.7 & 69.3$\pm$0.8 & \textbf{60.2}$\pm$1.6\\
    \hline
    \texttt{linearFGW}-Kmeans (Ours) & 98.7$\pm$1.2 & 70.58$\pm$0.57 & 71.46$\pm$1.03 & 54.49$\pm$0.3\\
    \texttt{linearFGW}-SC (Ours) & 98.2 $\pm$0.83 & \textbf{72.70}$\pm$0.03 & \textbf{73.33}$\pm$0.82 & 58.3$\pm$0.8\\
    \hline\hline
    \end{tabular}
    \label{tab:clusteringresults}
\end{table}

\textbf{Runtime Analysis:} By using the \texttt{linearFGW}, we can reduce the computational complexity of calculating the pairwise \texttt{FGW} distance for a data set of $N$ graphs from a quadratic complexity in $N$ i.e. $N(N-1)/2$) to linear complexity i.e. $N$ calculation of the \texttt{FGW} distances from graphs to the reference measure graph. We compare the running time of \texttt{linearFGW} and \texttt{FGW} with the same setting as in the classification task with fixed $\alpha$ of 0.5 and 0.0 for labeled graph data sets and IMDB-B (unlabeled), respectively. In Table \ref{tab:runningtimeanalysis}, we report the total running time of methods (both training time and inference time) on 5 data sets used for classification experiments . It is shown that the \texttt{linearFGW} is much faster than \texttt{FGW} on all considered data sets (roughly 7 times faster on COX2, BZR, ENZYMES, PROTEINS and 3 times faster on IMDB-B). These numbers confirm the computational efficiency of \texttt{linearFGW}, making it possible to analyze large-scale graph data sets.

\begin{table}[]
    \centering
    \caption{The total training time and inference time (in seconds) averaged over 10-folds of cross-validation (with fixed $\alpha$) for different data sets. The standard deviation is reported with the symbol $\pm$.}
    \begin{tabular}{l l l l l l}
    \hline\hline
    Methods/Data sets & COX2 & BZR & ENZYMES & PROTEINS & IMDB-B \\
    \hline
    \texttt{FGW} & 520.21$\pm$21.15 & 347.78$\pm$5.21 & 817.31$\pm$7.49 & 3224.36$\pm$125.02 & 1235.33$\pm$83.28\\
   \texttt{linearFGW} & 72.43 $\pm$ 0.16 & 53.81 $\pm$ 0.2 & 146.26 $\pm$1.64 & 431.25$\pm$9.25 & 358.92$\pm$10.41\\  
   \hline\hline
    \end{tabular}
    \label{tab:runningtimeanalysis}
\end{table}

\section{CONCLUSION AND FUTURE WORK}
We have developed an OT-based distance for learning with graph structured data. The key idea of this method is to embed node feature and topology of a graph into a linear tangent space, where the Euclidean distance between two embeddings of two graphs approximates their \texttt{FGW} distance. In fact the proposed distance is a generalization of the linear optimal transport \cite{wang2013linear} and the \texttt{FGW} distance. Thus it has the following advantages: 1) as the \texttt{FGW} distance, the proposed distance allows to take into account node features and topologies of graphs into the OT problem for computing the dissimilarity between two graphs,  2) we can derive a valid kernel from the proposed distance for graphs while the existing OT-based graph kernels are invalid, and 3) it provides the fast approximation of pairwise \texttt{FGW} distance, making it more efficient to deal with large-scale graph data sets. We conducted experiments on some benchmark graph data sets on both classification and clustering tasks, demonstrating the effectiveness of the proposed distance. 

In this work, we suggested to use the fused Gromov-Waserstein barycenter \cite{titouan2019optimal} as the reference measure graph. Thanks to the differentiablity of OT frameworks using techniques such as entropic regularization \cite{cuturi2013sinkhorn}, one possibility for future work is to learn the reference measure graph by updating the reference to minimize the supervision loss. The classification performance will be improved with the label information of graphs used in the training process. Another possibility would be to incorporate the \texttt{linearFGW} into graph-based deep learning models for learning with graph structured data.

\bibliographystyle{abbrvnat}
\bibliography{references}  






\appendix
\section{Appendix}
\subsection{Proof of Lemma \ref{lemma:optimalplan}}.
\begin{proof}


By the definition of barycentric projections for nodes and edges (see Definition \ref{def:barycentricprojections}), we denote $\Tilde{\mathbf{z}}_{k}=T_{\text{n},\pi^{*}}(\mathbf{z}_{k})$ and $\Tilde{\mathbf{C}}_{k,l}=T_{\text{e},\pi^{*}}(\mathbf{C}_{k,l})$ with $k,l=\overline{1,K}$. 
In order to prove the first claim, in contrary we assume that $\pi$ ($\neq\operatorname{diag}(\sigma)$) is the optimal transport plan with respect to $\texttt{FGW}_{\alpha}(\overline{\mathcal{G}}, \Tilde{\mathcal{G}})$. Then, we have the following inequality:
\begin{align}
\label{eqn:assumption}
   \sum_{k,l} (1-\alpha)\sigma_{k}\lVert \mathbf{z}_{k}-\Tilde{\mathbf{z}}_{k} \rVert^{2}+\alpha\sigma_{k}\sigma_{l}|\mathbf{C}_{k,l}-\Tilde{\mathbf{C}}_{k,l}|^{2} > \sum_{k,l,k^\prime,l^\prime}\left[(1-\alpha)\lVert \mathbf{z}_{k}-\Tilde{\mathbf{z}}_{k^\prime}\rVert^{2}+\alpha |\mathbf{C}_{k,l}-\Tilde{\mathbf{C}}_{k^\prime, l^\prime}|^{2}\right]\pi_{k,k\prime}\pi_{l,l\prime}
\end{align}

We rewrite the $\texttt{FGW}$ distance between $\overline{\mathcal{G}}(\mathbf{Z}, \mathbf{C}, \sigma)$ and $\mathcal{G}(\mathbf{X}, \mathbf{A}, \mu)$ as follows:
\begin{align*}
    \texttt{FGW}_{\alpha}(\overline{\mathcal{G}}, \mathcal{G}) =& \sum_{i,j,k,l}\left[(1-\alpha)\lVert \mathbf{x}_{i}-\mathbf{z}_{k}\rVert^{2}+\alpha \left(\mathbf{A}_{i,j}-\mathbf{C}_{k,l}\right)^{2}\right]\pi^{*}_{k,i}\pi^{*}_{l,j}\\
    =& (1-\alpha)\sum_{i}\mu_{i}\lVert \mathbf{x}_{i} \rVert^{2} + (1-\alpha)\sum_{k}\sigma_{k}\lVert \mathbf{z}_{k} \rVert^{2} - 2(1-\alpha)\sum_{i,k}\pi^{*}_{k,i}\langle \mathbf{x}_{i},\mathbf{z}_{k}\rangle\\
    +& \alpha \sum_{i,j}\mu_{i}\mu_{j}\mathbf{A}^{2}_{i,j}+ \alpha \sum_{k,l}\sigma_{k}\sigma_{l}\mathbf{C}_{k,l}^{2} -2\alpha \sum_{i,j,k,l}\mathbf{C}_{k,l}\mathbf{A}_{i,j}\pi^{*}_{k,i}\pi^{*}_{l,j}
\end{align*}

Using the definition of barycentric projections for nodes and edges in Definition \ref{def:barycentricprojections}, we have:
\begin{align*}
    \texttt{FGW}_{\alpha}(\overline{\mathcal{G}}, \mathcal{G}) =& (1-\alpha)\sum_{i}\mu_{i}\lVert \mathbf{x}_{i} \rVert^{2} + (1-\alpha)\sum_{k}\sigma_{k}\lVert \mathbf{z}_{k} \rVert^{2} - 2(1-\alpha)\sum_{k}\sigma_{k}\langle \mathbf{z}_{k},\Tilde{\mathbf{z}}_{k}\rangle\\
    +&\alpha \sum_{i,j}\mu_{i}\mu_{j}\mathbf{A}^{2}_{i,j}+ \alpha \sum_{k,l}\sigma_{k}\sigma_{l}\mathbf{C}_{k,l}^{2}-2\alpha \sum_{k,l}\sigma_{k}\sigma_{l}\mathbf{C}_{k,l}\Tilde{\mathbf{C}}_{k,l}\\
    =&(1-\alpha)\sum_{i}\mu_{i}\lVert \mathbf{x}_{i} \rVert^{2}-(1-\alpha)\sum_{k}\sigma_{k}\lVert\Tilde{\mathbf{z}}_{k}\rVert^{2}+(1-\alpha)\sum_{k}\sigma_{k}\lVert \mathbf{z}_{k}-\Tilde{\mathbf{z}}_{k} \rVert^{2}\\
    +& \alpha \sum_{i,j}\mu_{i}\mu_{j}\mathbf{A}_{i,j}^{2}-\alpha \sum_{k,l}\sigma_{k}\sigma_{l}\Tilde{\mathbf{C}}_{k,l}^{2}+\alpha \sum_{k,l}\sigma_{k}\sigma_{l}|\mathbf{C}_{k,l}-\Tilde{\mathbf{C}}_{k,l}|^{2}
    \label{eqn:2}
\end{align*}

By using the inequality (\ref{eqn:assumption}), we have:
\begin{align*}
\texttt{FGW}_{\alpha}(\overline{\mathcal{G}}, \mathcal{G}) >& (1-\alpha)\sum_{i}\mu_{i}\lVert \mathbf{x}_{i} \rVert^{2}-(1-\alpha)\sum_{k}\sigma_{k}\lVert\Tilde{\mathbf{z}}_{k}\rVert^{2}+\alpha \sum_{i,j}\mu_{i}\mu_{j}\mathbf{A}_{i,j}^{2}-\alpha \sum_{k,l}\sigma_{k}\sigma_{l}\Tilde{\mathbf{C}}_{k,l}^{2}\\
+& \sum_{k,l,k^\prime,l^\prime}\left[(1-\alpha)\lVert \mathbf{z}_{k}-\Tilde{\mathbf{z}}_{k^\prime}\rVert^{2}+\alpha |\mathbf{C}_{k,l}-\Tilde{\mathbf{C}}_{k^\prime, l^\prime}|^{2}\right]\pi_{k, k\prime}\pi_{l, l\prime}\\
=& \underbrace{(1-\alpha)\sum_{i}\mu_{i}\lVert \mathbf{x}_{i} \rVert^{2} -(1-\alpha)\sum_{k^\prime}\left(\sigma_{k^\prime}\lVert\Tilde{\mathbf{z}}_{k^\prime}\rVert^{2}-\sum_{k}\lVert \mathbf{z}_{k}-\Tilde{\mathbf{z}}_{k^\prime}\rVert^{2}\pi_{k, k^\prime}\right)}_{(a)}\\
+& \underbrace{\alpha \sum_{i,j}\mu_{i}\mu_{j}\mathbf{A}_{i,j}^{2} - \alpha \sum_{k^\prime, l^\prime}\left(\sigma_{k^\prime}\sigma_{l^\prime}\Tilde{\mathbf{C}}_{k^\prime, l^\prime}^{2} -\sum_{k,l}|\mathbf{C}_{k,l}-\Tilde{\mathbf{C}}_{k^\prime, l^\prime}|^{2}\pi_{k, k^\prime}\pi_{l, l^\prime}\right)}_{(b)}
\end{align*}

We first process the part (a) by using Jensen inequality as follows:
\begin{align*}
    (a) \geq & (1-\alpha)\sum_{i}\mu_{i}\lVert \mathbf{x}_{i} \rVert^{2}-(1-\alpha)\sum_{k^\prime, i}\frac{\pi^{*}_{k^\prime, i}}{\sigma_{k^\prime}}\left(\sigma_{k^\prime}\lVert \mathbf{x}_{i}\rVert^{2} - \sum_{k}\lVert \mathbf{z}_{k}-\mathbf{x}_{i}\rVert^{2}\pi_{k, k^\prime} \right)\\
    =& (1-\alpha)\sum_{k,k^\prime, i}\lVert \mathbf{z}_{k}-\mathbf{x}_{i}\rVert^{2}\frac{\pi_{k, k^\prime}\pi^{*}_{k^\prime, i}}{\sigma_{k^\prime}}=(1-\alpha)\sum_{k,i}\lVert  \mathbf{z}_{k}-\mathbf{x}_{i}\rVert^{2}\pi^{**}_{k,i}
\end{align*}
where $\pi^{**}_{k,i}=\sum_{k^\prime}\frac{\pi_{k, k^\prime}\pi^{*}_{k^\prime, i}}{\sigma_{k^\prime}}$ and $\pi^{**}$ is an admissible transport map from $\overline{\mathcal{G}}$ to $\mathcal{G}$.

We then process the part (b) in a similar way and obtain:
\begin{align*}
    (b) \geq & \alpha \sum_{i,j,k,l}|\mathbf{A}_{i,j}-\mathbf{C}_{k,l}|^{2}\pi^{**}_{k,i}\pi^{**}_{l,j}
\end{align*}
Combining parts (a) and (b), we have:
\begin{align*}
    \texttt{FGW}_{\alpha}(\overline{\mathcal{G}}, \mathcal{G}) > \sum_{i,j,k,l}\left[(1-\alpha)\lVert \mathbf{x}_{i}-\mathbf{z}_{k}\rVert^{2}+\alpha \left(\mathbf{A}_{i,j}-\mathbf{C}_{k,l}\right)^{2}\right]\pi^{**}_{k,i}\pi^{**}_{l,j}\\
\end{align*}
which contradicts the optimality of $\pi^{*}$. So we can conclude here that:
\begin{equation}
\label{eqn:optimal}
    \texttt{FGW}_{\alpha}(\overline{\mathcal{G}}, \Tilde{\mathcal{G}})=\sum_{k,l} (1-\alpha)\sigma_{k}\lVert \mathbf{z}_{k}-\Tilde{\mathbf{z}}_{k} \rVert^{2}+\alpha\sigma_{k}\sigma_{l}|\mathbf{C}_{k,l}-\Tilde{\mathbf{C}}_{k,l}|^{2}
\end{equation}

Proving the second claim is straight-forward by applying Jensen inequality for Equation (\ref{eqn:optimal}). Indeed, we have:
\begin{align*}
    \texttt{FGW}_{\alpha}(\overline{\mathcal{G}}, \Tilde{\mathcal{G}}) \leq & \sum_{k,l}(1-\alpha)\sigma_{k}\sum_{i}\frac{\pi^{*}_{k,i}}{\sigma_{k}}\lVert \mathbf{z}_{k}-\mathbf{x}_{i}\rVert^{2}+\alpha \sigma_{k}\sigma_{l}\sum_{i,j}\frac{\pi^{*}_{k,i}\pi^{*}_{l,j}}{\sigma_{k}\sigma_{l}}|\mathbf{C}_{k,l}-\mathbf{A}_{i,j}|^{2}\\
    =& \sum_{k,l,i,j}\left[(1-\alpha)\lVert \mathbf{z}_{k}-\mathbf{x}_{i}\rVert^{2}+\alpha |\mathbf{C}_{k,l}-\mathbf{A}_{i,j}|^{2}\right]\pi^{*}_{k,i}\pi^{*}_{l,j}=\texttt{FGW}_{\alpha}(\overline{\mathcal{G}}, \mathcal{G})
\end{align*}
\end{proof}

\subsection{Proof of Lemma \ref{Lemma2}}.
\begin{proof}
Let denote $\pi_{1}$ and $\pi_{2}$ be the optimal transport plans from $\overline{\mathcal{G}}$ to $\mathcal{G}_{1}$ and $\mathcal{G}_{2}$, respectively, in the sense of the \texttt{FGW} distance. We also denote $\Tilde{\mathcal{G}}_{1}$ and $\Tilde{\mathcal{G}}_{2}$ be the measure graphs which are transported from $\overline{\mathcal{G}}$ using the barycentric projections $\{T_{\text{n},\pi_{1}}, T_{\text{e},\pi_{1}}\}$ and $\{T_{\text{n},\pi_{2}}, T_{\text{e},\pi_{2}}\}$, respectively.

By using triangle inequality, we have:

\begin{align*}
    |\texttt{FGW}_{\alpha}(\mathcal{G}_{1}, \mathcal{G}_{2})-&\texttt{linearFGW}_{\alpha}(\mathcal{G}_{1},\mathcal{G}_{2})|\leq  |\texttt{FGW}_{\alpha}(\mathcal{G}_{1}, \mathcal{G}_{2})-2\texttt{FGW}_{\alpha}(\Tilde{\mathcal{G}}_{1},\mathcal{G}_{2})| + |2\texttt{FGW}_{\alpha}(\Tilde{\mathcal{G}}_{1},\mathcal{G}_{2})-\texttt{FGW}_{\alpha}(\Tilde{\mathcal{G}}_{1},\Tilde{\mathcal{G}}_{2})| \\
    +& |\texttt{FGW}_{\alpha}(\Tilde{\mathcal{G}}_{1},\Tilde{\mathcal{G}}_{2})-\texttt{linearFGW}_{\alpha}(\mathcal{G}_{1}, \mathcal{G}_{2})| \\
    \leq & 2 \texttt{FGW}_{\alpha}(\mathcal{G}_{1},\Tilde{\mathcal{G}}_{1}) + 2 \texttt{FGW}_{\alpha}(\mathcal{G}_{2},\Tilde{\mathcal{G}}_{2}) + \underbrace{|\texttt{FGW}_{\alpha}(\Tilde{\mathcal{G}}_{1}, \Tilde{\mathcal{G}}_{2})-\texttt{linearFGW}_{\alpha}(\mathcal{G}_{1}, \mathcal{G}_{2})|}_{(c)}
\end{align*}
The last inequality is obtained by using the relaxed triangle inequality of the \texttt{FGW} with $q=2$ (see \cite{titouan2019optimal}).
It is obvious to see that $\texttt{FGW}_{\alpha}(\mathcal{G}_{1},\Tilde{\mathcal{G}}_{1})\leq \texttt{diam}(\mathcal{G}_{1})$ and $\texttt{FGW}_{\alpha}(\mathcal{G}_{2},\Tilde{\mathcal{G}}_{2})\leq \texttt{diam}(\mathcal{G}_{2})$. In order to process the part (c), we notice that:
\begin{align}
\label{inequality:1}
\begin{split}
    |2\texttt{FGW}_{\alpha}(\Tilde{\mathcal{G}}_{1}, \overline{\mathcal{G}})-&\texttt{linearFGW}_{\alpha}(\mathcal{G}_{1},\mathcal{G}_{2})|= \lvert(1-\alpha) \sum_{k}2\sigma_{k}\lVert \mathbf{z}_{k} - T_{\text{n}, \pi_{1}}(\mathbf{z}_{k}) \rVert^{2}+\alpha\sum_{k,l}2\sigma_{k}\sigma_{l}|\mathbf{C}_{k,l}-T_{\text{e},\pi_{1}}(\mathbf{C}_{k,l})|^{2}\\
    -& (1-\alpha) \sum_{k}\sigma_{k}\lVert T_{\text{n},\pi_{1}}(\mathbf{z}_{k}) - T_{\text{n},\pi_{2}}(\mathbf{z}_{k}) \rVert^{2}-\alpha\sum_{k,l}\sigma_{k}\sigma_{l}|T_{\text{e},\pi_{1}}(\mathbf{C}_{k,l})-T_{\text{e},\pi_{2}}(\mathbf{C}_{k,l})|^{2}|\\
    \leq& (1-\alpha)\sum_{k}\sigma_{k}|2\lVert \mathbf{z}_{k} - T_{\text{n},\pi_{1}}(\mathbf{z}_{k}) \rVert^{2}-\lVert T_{\text{n},\pi_{1}}(\mathbf{z}_{k}) - T_{\text{n},\pi_{2}}(\mathbf{z}_{k}) \rVert^{2} | \\
    +& \alpha \sum_{k,l}\sigma_{k}\sigma_{l}|2|\mathbf{C}_{k,l}-T_{\text{e},\pi_{1}}(\mathbf{C}_{k,l})|^{2} - |T_{\text{e},\pi_{1}}(\mathbf{C}_{k,l})-T_{\text{e},\pi_{2}}(\mathbf{C}_{k,l})|^{2}|\\
    \leq & (1-\alpha)\sum_{k}2\sigma_{k}\lVert \mathbf{z}_{k} - T_{\text{n},\pi_{2}}(\mathbf{z}_{k}) \rVert^{2} + \alpha\sum_{k,l}2\sigma_{k}\sigma_{l}|\mathbf{C}_{kl}-T_{\text{e},\pi_{2}}(\mathbf{C}_{k,l})|^{2}\\
    =& 2\texttt{FGW}_{\alpha}(\Tilde{\mathcal{G}}_{2}, \overline{\mathcal{G}})
\end{split}
\end{align}
The last inequality is obtained by applying the following inequality: $|2(a-b)^{2}-(b-c)^{2}|\leq (a-c)^{2}$, for all $a,b,c\in \mathbb{R}$. Finally, we have:
\begin{align*}
    (\text{c})=|\texttt{FGW}_{\alpha}(\Tilde{\mathcal{G}}_{1}, \Tilde{\mathcal{G}}_{2})-\texttt{linearFGW}_{\alpha}(\mathcal{G}_{1}, \mathcal{G}_{2})|\leq & |\texttt{FGW}_{\alpha}(\Tilde{\mathcal{G}}_{1}, \Tilde{\mathcal{G}}_{2})-2\texttt{FGW}_{\alpha}(\Tilde{\mathcal{G}}_{1}, \overline{\mathcal{G}})|+|2\texttt{FGW}_{\alpha}(\Tilde{\mathcal{G}}_{1}, \overline{\mathcal{G}})-\texttt{linearFGW}_{\alpha}(\mathcal{G}_{1}, \mathcal{G}_{2})|\\
    \leq & 2 \texttt{FGW}_{\alpha}(\Tilde{\mathcal{G}}_{2}, \overline{\mathcal{G}}) + 2\texttt{FGW}_{\alpha}(\Tilde{\mathcal{G}}_{2}, \overline{\mathcal{G}})=4 \texttt{FGW}_{\alpha}(\Tilde{\mathcal{G}}_{2}, \overline{\mathcal{G}})\\
    \leq & 4\texttt{FGW}_{\alpha}(\mathcal{G}_{2}, \overline{\mathcal{G}})
\end{align*}
The second inequality is obtained by applying the relaxed triangle inequality of the \texttt{FGW} with $q=2$ (see \cite{titouan2019optimal}) and the inequality (\ref{inequality:1}) while the last inequality is obtained by the second claim of Lemma \ref{lemma:optimalplan}. We also have $(\text{c})\leq 4\texttt{FGW}_{\alpha}(\mathcal{G}_{1}, \overline{\mathcal{G}})$ by the symmetry of $\mathcal{G}_{1}$ and $\mathcal{G}_{2}$ with respect to $\overline{\mathcal{G}}$, which concludes the proof.
\end{proof}
\end{document}